# Ask Me Even More: Dynamic Memory Tensor Networks (Extended Model)

**Govardana Sachithanandam Ramachandran**  **Ajay Sohmshetty**
Department of Computer Science   Department of Computer Science
Stanford University   Stanford University
rgsachin@{gmail.com or stanford.edu}   ajay14@stanford.edu

## Abstract

We examine Memory Networks for the task of question answering (QA), under common real world scenario where training examples are scarce and under weakly supervised scenario, that is only extrinsic labels are available for training. We propose extensions for the Dynamic Memory Network (DMN), specifically within the attention mechanism, we call the resulting Neural Architecture as Dynamic Memory Tensor Network (DMTN). Ultimately, we see that our proposed extensions results in over 80% improvement in the number of task passed against the baselined standard DMN and 20% more task passed compared to state-of-the-art End-to-End Memory Network for Facebook's single task weakly trained 1K bAbI dataset.

## 1   Introduction

Wielding the ability to answer open ended questions, an effective question answering (QA) system can be incredibly powerful, as many tasks in natural language processing can be modelled into QA problems. Currently, several state-of-the-art memory networks exist for QA. Specifically, Dynamic Memory Networks (DMNs), End-to-End Neural Networks among others, have all been applied to this task all with reasonable degrees of success. We will focus on the application of DMNs to this task, while relegating End-to-End Neural Networks as a baseline exploration. We implement and extend the DMN architecture as described in the sections below.

## 2   Datasets

Given a series of input statements, interspersed with open-ended questions, our task is to produce an answer corresponding to each question at every respective point in time.

### 2.1   Facebook's babI Dataset

We are using Facebook's babI dataset for this task. This dataset is comprised of 20 different reading comprehension tasks, shown in Table 1, designed to measure understanding [5]. For each task, we have 1k training examples; the following is an example data point for the "Two Supporting Facts" task:

1 Mary got the milk there.
2 John moved to the bedroom.
3 Sandra went back to the kitchen.
4 Mary travelled to the hallway.
5 Where is the milk?         hallway   1 4

| # | Tasks |
|---|---|
| 1 | single-supporting-fact |
| 2 | two-supporting-facts |
| 3 | three-supporting-facts |
| 4 | two-arg-relations |
| 5 | three-arg-relations |
| 6 | yes-no-questions |
| 7 | counting |
| 8 | lists-sets |
| 9 | simple-negation |
| 10 | indefinite-knowledge |
| 11 | basic-coreference |
| 12 | conjunction |
| 13 | compound-coreference |
| 14 | time-reasoning |
| 15 | basic-deduction |
| 16 | basic-induction |
| 17 | positional-reasoning |
| 18 | size-reasoning |
| 19 | path-finding |
| 20 | agents-motivations |

Table:1

6 John got the football there.
7 John went to the hallway.
8 Where is the football?    hallway  6 7

Perhaps the most interesting artifact about this dataset is that it is a synthetic dataset; it is machine generated. Therefore, it is subject to overfitting, and generalizability may be of concern, though it is a sizeable dataset.

## 3    Related Works

### 3.1    End-To-End Memory Networks

Memory networks (MemNN) by Weston, Jason et al[6] introduces the concept of using long-term memory component and inference components for reasoning tasks. An extension to this model is End-To-End Memory Networks MemN2N) by Sukhbaatar, S et al[7], it extends MemNN to be trained end-to-end, that is, it is weakly supervised. The model uses attention mechanism and cycles over the inputs with multiple computational steps -- or "hops" -- before it outputs answer. The model is continuous hence making it trainable end-end by back propagation. The model has following three modules.

**Input Memory Module:** Given a set of inputs $x_1 ... x_n$, a transformation matrix $A$ is used to convert each of the inputs to corresponding distributed representation $\{m\}$ of dimension $d$. Similar transformation is done for the query $q$ which is transformed to $u$ by another matrix $B$. In the embedded space input $u$ and query $u$ are matched using softmax

$$p_i = Softmax(u^T m_i)$$

**Output Memory Module:** The inputs have an output vector representation as well $c_i$ generated by another embedding matrix $C$. The output memory vector is given by

$$o = \sum_i p_i c_i$$

where $p_i$ is the probability vector from the input.

**Prediction:** The output labels are predicted by a softmax on the sum of output memory vector and query vector

$$\hat{a} = Softmax(W(o + u))$$

These 3 modules, constitutes a single hop on the input data, multiple hops are achieved by passing sum of input and output memory vector as in the input vector for the subsequent hop

$$u^{k+1} = u^k + o^k$$

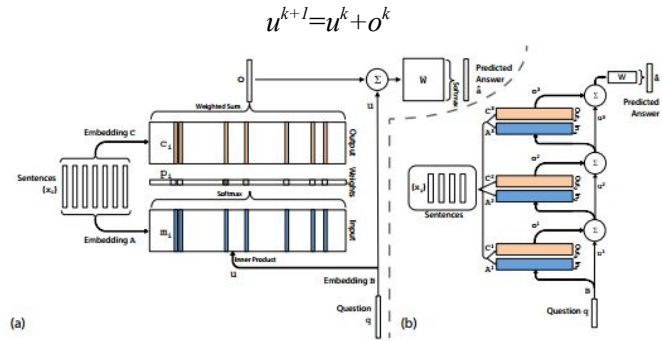

**Figure 1**: (a): A single layer version of MemN2N. (b): Three hops variant of MemN2N

### 3.2    Dynamic Memory Networks (DMNs)

Dynamic Memory Networks, introduced by Ankit Kumar et al., contain four main modules: an input module, a question module, an episodic memory module; and finally an answer module [3]. We provide a brief overview of the main modules and redirect the reader to the full paper for the full details [3].

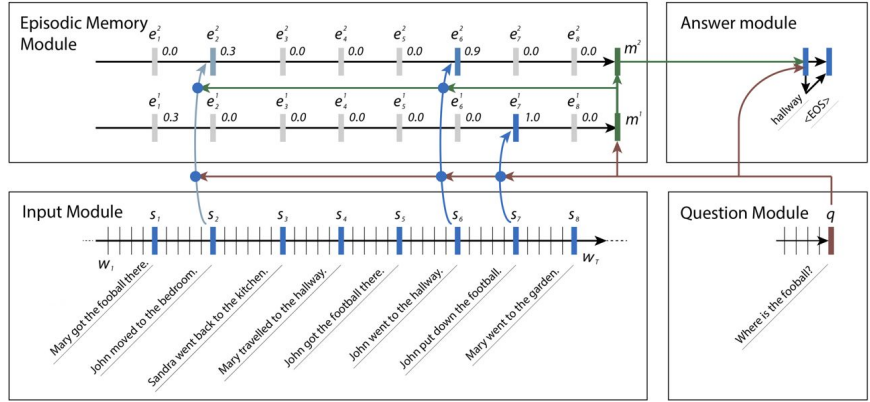

**Figure 2**: Overview of DMN

**Input Module:** The input module is designed to encode inputs into distributed word vector representations. Given a sequence of words, where each word is converted into vector representations using GloVe, we encode the input sequence through a gated recurrent network (GRU). For cases where there are multiple sentences in the input, we insert end-of-sentence tokens after each sentence, allowing the network to process lists of sentences. We then store each memory, through vector representations, for future reference by the episodic memory module.

**Question Module**: The question module also encodes question inputs into vector representations using the same GRU RNN as in the input module, but instead for feeding as the initial state into the episodic memory module.

**Episodic Memory Module:** The episodic memory module, through an attention mechanism identifies memories that are relevant and iterates through stored inputs to form an answer vector. This module receives the question vector representation from the question module, and sets that as the initial state into a modified GRU RNN.

**Answer Module:** The answer module converts the final state produced by the episodic memory module into a tokenized answer, using, again, a GRU RNN.

## 4    Dynamic Memory Tensor Networks - Extended Model

### 4.1    Attention Mechanism

The current DMN uses a gating function for attention mechanism. At each pass *i*, the mechanism takes as input a candidate fact $c_t$, a previous memory $m^{i-1}$, and the question *q* to compute a gate:

$$g_t^i = G(c_t, m^{i-1}, q)$$

where, the scoring function G takes a feature vector of handcrafted similarity measures, and returns a gate score:

$$z(c, m, q) = [c, m, q, c \circ q, c \circ m, |c - q|, |c - m|, c^T W^{(b)} q, c^T W^{(b)} m]$$
$$G(c, m, q) = \sigma(W^{(2)} tanh(W^{(1)} z(c, m, q) + b^{(1)}) + b^{(2)})$$

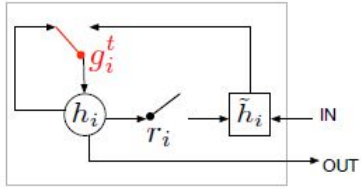

Figure 3: Attention Mechanism for DMN

The scoring function is used as gate retention weights for the computing the episode of each pass of the input. Where episode in each pass is represented as

$$e^i = h^i_{T_C}$$
$$h^i_t = g^i_t GRU(c_{t,}, h^i_{t-1}) + (1 - g^i_t) h^i_{t-1}$$

**Neural Tensor Network (NTN) for Attention Mechanism:** Form the neural architecture, it is clear that attention Mechanism is key for the performance of the system. Currently handcrafted similarity measure are used as feature vector for scoring, we propose mechanism to let the Neural Network craft the similarity measure, We propose the use of Neural Tensor Network [9] for this purpose. The scoring function now will be of the form

$$G(c, m, q) = \sigma(W^{(2)} tanh((c^T W_R^{[1:k]} q) m^T + c^T W_{cq}^{[1:k]} q + m^T W_{mq}^{[1:k]} q + c^T W_{cm}^{[1:k]} m + V_R [c\ q\ m]^T + b_R) + b^{(2)})$$

Where, $W_R^{[1:k]} \varepsilon \mathbb{R}^{d \times d \times d \times k}$ represent the three way relationship tensor between $(c, m, q)$. $W_{cq}^{[1:k]}, W_{cm}^{[1:k]}, W_{mq}^{[1:k]} \varepsilon \mathbb{R}^{d \times d \times k}$ represents two way relationship tensor between a pair of entries $\in \{c, m, q\}$. The other relationship $R$, parameters are $V_R \varepsilon \mathbb{R}^{k \times 3d}$ and $b_R \varepsilon \mathbb{R}^k$.

In our experiment, we found having three way relationship helps but due to constraints of resources (GPU) and time taken for convergence and hyper parameter tuning -- dropout on weights $W_R^{[1:k]}$ -- and regularization, In our final experiments we have dropped three way relationship, that is reducing the scoring function to:

$$G(c, m, q) = \sigma(W^{(2)} tanh(c^T W_{cq}^{[1:k]} q + m^T W_{mq}^{[1:k]} q + c^T W_{cm}^{[1:k]} m + V_R [c\ q\ m]^T + b_R) + b^{(2)})$$

### 4.1.1 Encapsulating Models

NTN covers wide range of similarity score, Given entities $e_1$ and $e_2$, if there exist $R$ relationship between them the NTN scoring function $g(e_1 R e_2)$ covers the following similarity functions.

**Distance Model:** The distance cores builds relationship by mapping the left and right entities to a common space using a relationship specific mapping matrix and measuring the L1 distance between the two. The scoring function

$$g(e_1, R, e_2) = \|W_{R,1} e_1 - W_{R,2} e_2\|_1$$

$W_{R,1}, W_{R,2} \varepsilon \mathbb{R}^{d \times d \times d}$ are the parameters of the relation $R$.

**Single Layer Model**: Compare to the first function, single layer neural network adds nonlinearity. The scoring function has the following form:

$$g(e_1, R, e_2) = u_R^T f(W_{R,1} e_1 + W_{R,2} e_2) = u_R^T f\left([W_{R,1} W_{R,2}] \begin{bmatrix} e_1 \\ e_2 \end{bmatrix}\right)$$

where, $f = tanh$, $W_{R,1}, W_{R,2} \varepsilon \mathbb{R}^{d \times d \times d}$ and $b_1, b_2 \varepsilon \mathbb{R}^{d \times 1}$ are the parameters of the relation.

**Hadamard Model:** The model tackles the issue of weak entity vector interaction through multiple

matrix products followed by Hadamard products,The scoring function has the following form:

$$g(e_1, R, e_2) = (W_1 e_1 \otimes W_{rel,1} e_R + b_1)^T (W_2 e_2 \otimes W_{rel,2} e_R + b_2)$$

where $W_1, W_{rel,1}, W_2, W_{rel,2} \in \mathbb{R}^{d \times d}$ and $b_1, b_2 \in \mathbb{R}^{d \times 1}$

**Bilinear Model:** Captures weak entity vector interaction through a relation-specific bilinear form. The scoring function is as follows:

$$g(e_1, R, e_2) = e_1^T W_R e_2$$

where $W_R \in \mathbb{R}^{d \times d}$, the parameter of relation $R$'s scoring function.

### 4.1.2 Extended - Neural Tensor Network

In order to increase the expressibility, we experimented with another form scoring function with promising results, here we define *z=[c,m,q],* and attention gate scoring function is of the form

$$G(c, m, q) = \sigma(W^{(2)} \tanh(z^T W_R^{[1:k]} z + V_R z^T + b_R) + b^{(2)})$$

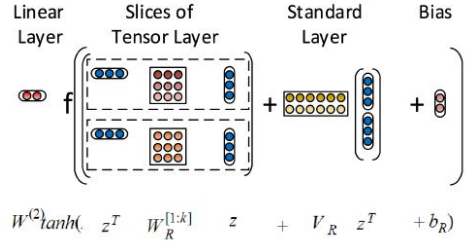

**Figure 4**: Illustration of Extended-NTN

Here $W_R^{[1:k]} \in \mathbb{R}^{3d \times 3d \times k}$, and we see the number of parameter reduced without losing the covered similarity function.

## 5 Experiments

### 5.1 Metrics

As done in other models in the Q&A problem space – MemN2N, DMN, etc.– we use accuracy for gauging the performance of the model. As done in MemN2N [7] and DMN [3] we define a task as passed when test accuracy is $\geq 95\%$.

### 5.2 Results

We weakly trained (i.e. without supporting facts feedback) DMN and DMTN on the 1K bAbi dataset. To measure the lift from DMN to DMTN, we chose the same hyperparameters across all of our compared tasks to be the same and between DMN and DMTN. In our reported results, due to constraints on resources (GPU) and time taken for convergence and hyper parameter tuning we restrict to our results to 2-way tensor relationship. For simplicity in our final bAbi experiment results, we used feed forward NN as answer module. MemN2N with Positional Encoding (PE) + Linear Start (LS) + Random empty memory (RN) is the current state-of-the-art (SOTA) for a weakly-trained single bAbi task [7] and we compare number of task passed by DMTN against the SOTA.

| DMTN | |
|---|---|
| Hyper-parameter | Value |
| Hidden Layer Dimension | 40 |
| Tensor Relationship Slice | 40 |
| Memory Hops | 5 |
| Random Dropout | 0 |
| Embedding size | 50 |
| Epoch | 150 |
| L2 | 0.0001 / 0.0003 |

**Table 2**: Hyperparameters used for DMTN.

| | | Single task weakly trained (1K data) | | | | All task joint - weakly trained (20 x 1K data) |
|---|---|---|---|---|---|---|
| | | | | | Published SOTA | |
| # | Tasks | DMN best* | DMN baseline | DMTN | MemN2N with PE,LS,RN | MemN2N with PE,LS,Lw |
| 1 | single-supporting-fact | 100 | 100 | 100 | 100 | 99.9 |
| 2 | two-supporting-facts | 32.7 | 29.9 | 35.9 | 91.7 | 81.2 |
| 3 | three-supporting-facts | 26.3 | 32 | 36 | 59.7 | 68.3 |
| 4 | two-arg-relations | 89.8 | 82.1 | 100 | 97.2 | 82.5 |
| 5 | three-arg-relations | 97.5 | 97.6 | 97.3 | 86.9 | 87.1 |
| 6 | yes-no-questions | 96.3 | 96.6 | 96.5 | 92.4 | 98 |
| 7 | counting | 80.3 | 77.6 | 80.1 | 82.7 | 89.9 |
| 8 | lists-sets | 76.6 | 98.6 | 98.3 | 90 | 93.9 |
| 9 | simple-negation | 94.1 | 95 | 95.3 | 86.8 | 98.5 |
| 10 | indefinite-knowledge | 95 | 90.2 | 93.2 | 84.9 | 97.4 |
| 11 | basic-coreference | 100 | 71.4 | 99.9 | 99.1 | 96.7 |
| 12 | conjunction | 100 | 67.2 | 99.8 | 99.8 | 100 |
| 13 | compound-coreference | 94.4 | 92.5 | 94.4 | 99.6 | 99.5 |
| 14 | time-reasoning | 77.6 | 74.7 | 73 | 98.3 | 98 |
| 15 | basic-deduction | 67 | 50.5 | 99.9 | 100 | 98.2 |
| 16 | basic-induction | 48.1 | 44.8 | 46.4 | 98.7 | 49 |
| 17 | positional-reasoning | 65.9 | 52 | 62.7 | 49 | 57.4 |
| 18 | size-reasoning | 93.7 | 91.5 | 95.8 | 88.9 | 90.8 |
| 19 | path-finding | 10.7 | 8.2 | 9.9 | 17.2 | 9.4 |
| 20 | agents-motivations | 98.1 | 97.6 | 98.2 | 100 | 99.8 |
| | passed task (acc>=95%) | 7 | 6 | 11 | 9 | 10 |

**Table 3**: Accuracies across all tasks for MemN2N, DMN, and DMTN. Here *DMN baselines* serves as the baseline for DTMN to measure the lift with the proposed changes. *DMN best\** is the best document performance of DMN with optimal hyperparameter tuning on bAbi weakly trained dataset- http://yerevann.github.io/2016/02/05/implementing-dynamic-memory-networks .

### 5.3 Analysis

Overall the performance of the DMTN is considerably better than standard DMN in all the tasks. Note that we didn't perform any hyperparamater tuning as against the baselined standard DMN. DMTN passes 80% more number of tasks than the baseline DMN . DMTN also clearly outperforms -- in number of tasks passed --  the SOTA MemN2N-PE,LS,RN both single task trained (1K) and jointly trained (20 x 1K) variants. Compared to MemN2N-PE,LS,RN model, DMTN passes 20% more tasks without any hyperparamater tuning.

For the tasks DMTN fails, we believe that they are limitation of other modules of DMN. Ex: Tasks 2, 3 performance are curtailed by the limited positional embedding, which plays significant role in SOTA [7].

### 5.4 Implementation

A Theano based DMTN reference implementation is at https://github.com/rgsachin/DMTN

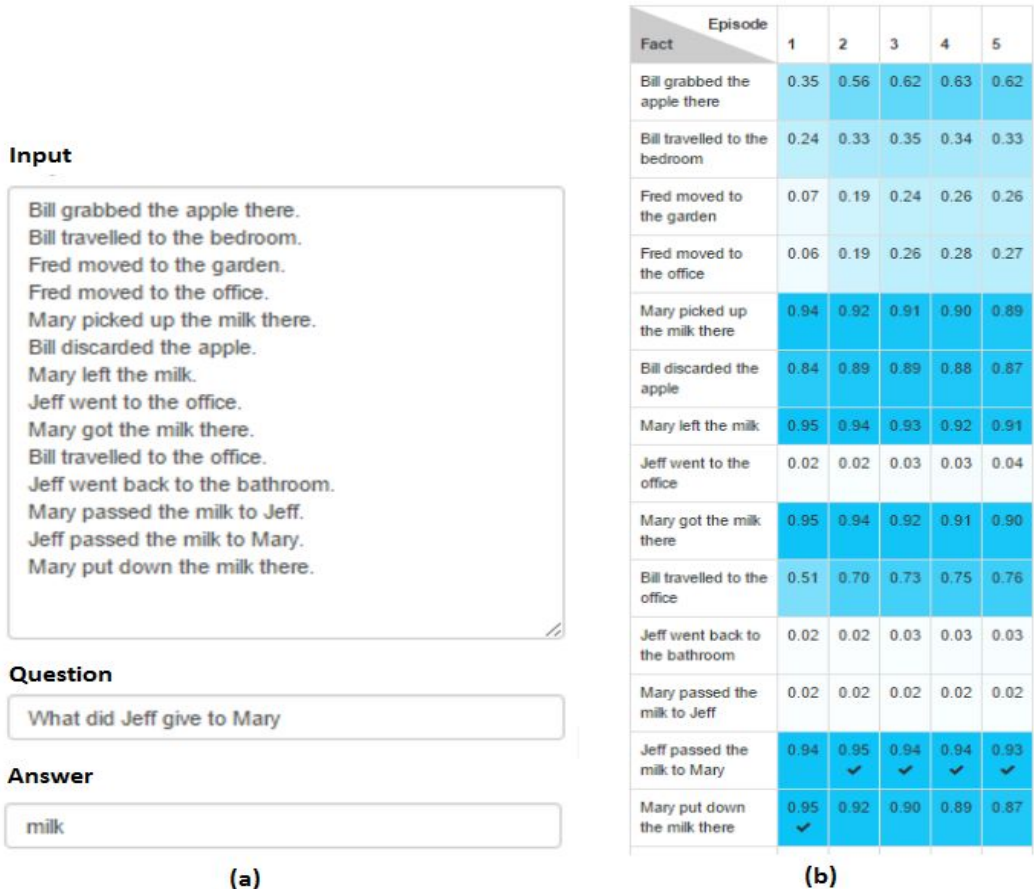

**Figure 6**: (a): Is the input, question and the answer return by DMTN model. (b) is heat map of the gate weight on the attentions mechanism for each of the input fact for each of five hops.

## 6 Conclusion and Future Work

Under weakly supervised environment, DMTN clearly outperforms  the state-of-the-art for weakly trained 1K bAbi dataset, which is memN2N with PE,LS,RN by 20% in the number of tasks passed, as well as show clear and significant lift in the order of over 80% compared to the standard DMN results on the bAbi 1K dataset in the number of tasks passed.

**Future work**: For facts based tasks, with longer sentences, use of doc2vec for positional embedding(PE), initial experiments found to improve the accuracy of the model, especially for tasks 2 and 3. There is also a lot of scope for hyperparameter tuning to further outperform standard DMN and SOTA-MemN2N. Adding 3-way relationship tensor with dropout on Tensor weights $W_R^{[1:k]}$ showed considerable promise.